\documentclass[10pt,twocolumn,letterpaper]{article}

\usepackage{3dv}
\usepackage{times}
\usepackage{epsfig}
\usepackage{graphicx}
\usepackage{amsmath}
\usepackage{amssymb}
\usepackage{multirow}
\usepackage{adjustbox}
\usepackage{booktabs}
\usepackage{bbding}
\usepackage{subcaption}
\usepackage[font=small]{caption}
\usepackage{tabularx}
\usepackage{array}

\newcommand{\R}{\mathbb{R}}

\usepackage[pagebackref=true,breaklinks=true,colorlinks,bookmarks=false]{hyperref}

\threedvfinalcopy 

\def\threedvPaperID{268} 

\ifthreedvfinal\fi
\begin{document}

\title{
Occlusion Guided Self-supervised Scene Flow Estimation on 3D Point Clouds
}

\author{Bojun Ouyang\\
Tel Aviv University\\
{\tt\small bojungouyang@mail.tau.ac.il}
\and
Dan Raviv\\
Tel Aviv University\\
{\tt\small darav@tauex.tau.ac.il}
}

\maketitle
\thispagestyle{empty}

\begin{abstract}

Understanding the flow in 3D space of sparsely sampled points between two consecutive time frames is the core stone of modern geometric-driven systems such as VR/AR, Robotics, and Autonomous driving. The lack of real, non-simulated, labeled data for this task emphasizes the importance of self- or un-supervised deep architectures. This work presents a new self-supervised training method and an architecture for the 3D scene flow estimation under occlusions. Here we show that smart multi-layer fusion between flow prediction and occlusion detection outperforms traditional architectures by a large margin for occluded and non-occluded scenarios. We report state-of-the-art results on Flyingthings3D and KITTI datasets for both the supervised and self-supervised training.
\footnote{Our code will be publicly available upon publication.} 
\footnote{https://github.com/BillOuyang/3D-OGFlow.git}

\end{abstract}

\section{Introduction}

Due to the development of autonomous driving, robotic manufacturing, and virtual and augmented technologies, understanding the motion in the dynamic scene becomes important and critical in many backbones~\cite{menze_geiger_2015,BTSSPP15}. 
Unlike Optical Flow~\cite{7410673}, where we search for the projected 2D motion in the image, in Scene Flow~\cite{1388274}, we wish to find also the flow along the depth dimension. Traditionally, the scene flow estimation was performed on the stereo~\cite{5539791,4409000} or from RGB-D~\cite{8374580,6126509} sensors for indoor environments and used Light Detecting and Ranging (LiDAR) sensors in the outdoor scenes.

Switching from 2D to 3D introduces interesting challenges. While RGB images contain color information of the scene and are provided as a dense regular grid, the point clouds carry the geometric information and are presented as a sparse unordered set of points in space, which forces us to switch from traditional image-based algorithms to graph models. \emph{Axiomatic} methods, such as~\cite{10.1117/12.57955,chen_medioni}, found the rigid alignment between the point clouds by solving an energy minimization problem. Later, ~\cite{4270190} relaxed the rigid assumptions, but their optimization problem is hard to solve.



Moving from \emph{Axiomatic} models towards \emph{Learnable} architectures~\cite{liu:2019:flownet3d,gu_wang_wu_lee_wang_2019,puy20flot,wu2020pointpwc}, we have seen a large boost in performance and running time in supervised architectures.
Due to the lack of the annotated labels, there is a demand for self-supervised training methods for the scene flow estimation on point clouds. Among popular methods one can find,  ~\cite{wu2020pointpwc,Mittal_2020_CVPR} suggested minimizing the nearest neighbor distances between the target and the warped source according to the estimated flow,~\cite{DBLP:conf/3dim/ZuanazziVBM20} proposed self-supervised learning based on the adversarial metric learning techniques.

When we estimate the flow, we always encounter occlusion, where some regions in one scene might not exist in the other. This is mainly caused by the motion in the scene, so that some objects may enter or leave the visible zone of the camera. The main difficulty in flow estimation under occlusion relates to the connection between the flow correlation and the magnitude of the occlusion. On the one hand, given the occluded parts, we can optimize for the best flow, and on the other hand, given the best flow, we can conclude which part is occluded. In practice, optimizing those two unknowns in parallel is non-trivial in a self-supervised scheme due to possible collapse towards an all-occluded solution.


Although we already have extensive studies of occlusion in flow estimation in 2D images~\cite{Hur:2019:IRR, Zhao_2020_CVPR,saxena2019pwoc}, it is still an open challenge for 3D point clouds that merely no one has studied. Due to the difference in the information carried by these two data structures, directly utilizing those image-based occlusion handling techniques to the point cloud data does not provide the boost we need.~\cite{Ouyang_2021_CVPR} was the first to estimate the occlusion in point clouds, but their training method requires the ground truth occlusion label, which is almost impossible to acquire in the real scenario.


In this paper, we focus on the scene flow estimation problem on point clouds with occlusion. We present a self-supervised architecture called 3D-OGFlow that merges two networks across all layers, where one learns the flow and the other learns the occlusions. We further present a novel Cost Volume layer that can encode the similarity between the point clouds with occlusion handling. 
We show state-of-the-art performance on Flyingthings3D and KITTI scene flow 2015 benchmark for both occluded and non-occluded versions.

\section{Related Work}

\noindent
\textbf{Scene Flow Estimation on Point Clouds}.
Due to the increasing popularity of range data and the development of the 3D deep learning~\cite{liu2019pvcnn,allen_2010,qi_su_niebner_dai_yan_guibas_2016,wu_song_khosla_yu_zhang_tang_xiao_2015,charles_su_kaichun_guibas_2017,qi2017pointnetplusplus,Wang2019_GACNet,wu_qi_fuxin_2019}, many works such as~\cite{9320433,7759282,behl_paschalidou_donne_geiger_2019, rishav2020deeplidarflow,7989666,wang_suo_ma_pokrovsky_urtasun_2018,gu_wang_wu_lee_wang_2019,Kittenplon_2021_CVPR} suggested directly estimating the scene flow on the point clouds obtained from LiDAR scans. Based on the hierarchical architecture of~\cite{qi2017pointnetplusplus}, FlowNet3D~\cite{liu:2019:flownet3d} was the first to propose the flow embedding layer which can aggregate the features across consecutive frames. Inspired by the feature pyramid structure of~\cite{Sun2018:Model:Training:Flow}, PointPWC~\cite{wu2020pointpwc} suggested estimating the scene flow on multiple levels. They also introduced a novel Cost Volume that can aggregate the Matching Cost between the point clouds in a patch-to-patch manner using a learnable weighted sum. Later, FLOT~\cite{puy20flot} proposed a network that learns the correlation in an all-to-all manner based on the graph matching and optimal transport. 


\noindent
\textbf{Occlusion in Flow Estimation}.
In the optical flow or scene flow, handling the occlusion in images is important as it can highly influence the estimation accuracy. Many works in optical flow~\cite{Zhao_2020_CVPR,Hur:2019:IRR,Janai2018ECCV}, scene flow~\cite{saxena2019pwoc}, or both~\cite{ilg_saikia_keuper_brox_2018}, suggested using a CNN to learn the occlusion, which is further used to refine the predicted flow. Other works~\cite{hur_roth_2017,Wang_2018_CVPR,Meister:2018:UUL,Liu:2019:DDFlow,Liu:2019:SelFlow,Hur:2020:SSM} suggested estimating the occlusion using forward-backward consistency check. In~\cite{Zhao_2020_CVPR,saxena2019pwoc}, they excluded the occluded regions before the correlation/Cost Volume construction, which significantly improved the performance. When it comes to the point cloud data, ~\cite{Ouyang_2021_CVPR} recently suggested excluding the computed Cost Volume for the occluded points, but this can harm the flow estimation accuracy for the occluded regions.

Unlike~\cite{Zhao_2020_CVPR,saxena2019pwoc,Ouyang_2021_CVPR}, our method does not exclude the occluded regions during the correlation construction as they contain useful geometric information. Instead, We use a separate construction of the Cost Volume for the occluded and non-occluded regions according to their properties, and then we aggregate the two in an occlusion-weighted manner.

\noindent
\textbf{Self-supervised Learning}.
In the case of 2D images, minimizing the photometric loss between reference and warped target images is common in unsupervised learning.~\cite{Meister:2018:UUL,Liu:2019:DDFlow,Liu:2019:SelFlow,liu2020learning,Hur:2020:SSM} suggested excluding the occluded pixels in the photometric loss, which makes a lot more sense.~\cite{Liu:2019:DDFlow,Liu:2019:SelFlow} learned the optical flow for the occluded regions using data augmentation, while~\cite{liu2020learning} suggested making another forward inference on the augmented data as a regularization. When it comes to scene flow estimation on point clouds,~\cite{Mittal_2020_CVPR} take the supervised pretrained FlowNet3D~\cite{liu:2019:flownet3d} as a backbone and fine-tune it on the unannotated KITTI using the self-supervised nearest neighbor loss and cyclic consistency.~\cite{wu2020pointpwc} use the Chamfer distance together with the smoothness and Laplacian regularization as their self-supervised training losses. Recently,~\cite{DBLP:conf/3dim/ZuanazziVBM20} proposed a self-supervised training scheme based on metric learning. They use triplet loss and cyclic consistency and showed a remarkable results on the occluded Flyingthings3D~\cite{mayer_ilg_hausser_fischer_cremers_dosovitskiy_brox_2016} and KITTI~\cite{menze_geiger_2015,menze_heipke_geiger_2015}.

In our work, we suggest a novel self-supervised training scheme that can learn the scene flow for the occluded scene. Our strategy shows a significant improvement in the occluded datasets compared to the previous state-of-the-art.

\section{Problem Definition}
Consider the sampling of a 3D scene at two different time frames, denote $S=\{s_i\in \R^3\}_{i=1}^{n_1}$ the source sampling with $n_1$ points, and $T=\{t_j\in \R^3\}_{j=1}^{n_2}$ the target sampling with $n_2$ points. In addition to the $s_i$, $t_j$ that describe the spatial coordinate, each source and target point can also have an associated feature such as surface normal, which is denoted by $c_i\in\R^d$ and $g_j\in\R^d$ respectively.

For the scene flow estimation on point clouds, the goal is to find a 3D non-rigid flow $f(s_i)\in \R^3$ for every source point $s_i$ towards $T$, such that the warped source $S_w=\{s_i+f(s_i)\}_{i=1}^{n_1}$ has the best alignment with $T$. Due to the possible occlusion and the difference in the sampling, some points in $S$ might not exit in $T$. For this reason, we learn a flow representation for each $s_i\in S$ towards the $T$ instead of the correspondence between $S$ and $T$. 

We also want to find the occlusion label for every source point $s_i$ with respect to $T$, denoted by $O(s_i)$. $O(s_i)=1$ when $s_i$ is \textit{non-occluded}. When $O(s_i)=0$, it means $s_i$ is $occluded$, in other words it does not exist in the target frame.


\begin{figure*}
\begin{center}
\includegraphics[width=1.0\textwidth]{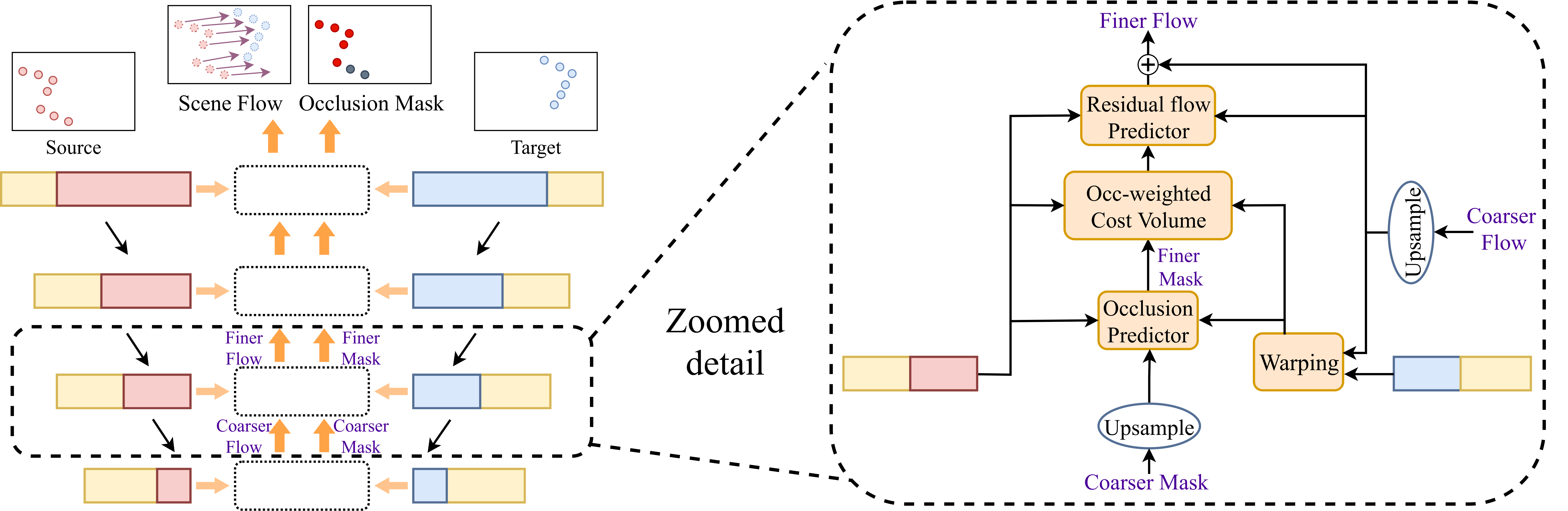}
\end{center}
\vspace{-20pt}
\caption{\textbf{3D-OGFlow}.On the left, we show our model's general structure. We use a feature pyramid structure to perform the downsampling of the point clouds and the feature encoding. On the right, we show the structure at each pyramid level. We first warp the target point cloud towards the source using the upsampled flow from the previous level. Then we construct our Occlusion-weighted Cost Volume using the predicted occlusion. Finally, we estimate the residual flow and add it to the upsampled flow to generate the finer scene flow.}
\label{fig:full_arch}
\end{figure*}

\section{Architecture}
The architecture of 3D-OGFlow is shown in Fig.\ref{fig:full_arch}. The inputs of the model are the source and the target point clouds sampled at different time frames. The outputs are the predicted scene flow $f(s_i)$ and occlusion label $O(s_i)$ for $s_i\in S$ with respect to $T$. We adopt the 4-level feature pyramid network in~\cite{wu2020pointpwc}, where we first generate the downsampled source ($S^l$) and target ($T^l$) point clouds for each pyramid level $l$ using Farthest Point Sampling (FPS)~\cite{qi2017pointnetplusplus}. Then, we use PointConv~\cite{wu_qi_fuxin_2019} to perform the convolution on the point clouds, which generates and increases the depth of the encoded features for the downsampled point clouds along the pyramid. At each pyramid level $l$, we first perform a backward warping from the target point cloud towards the source by using the upsampled flow from pyramid level $l+1$. The warping brings the two point clouds closer to each other such that the tracking of large motion can be more accurate. Second, we estimate the occlusion label for each source point using the occlusion predictor. Third, we construct our occlusion-weighted Cost Volume, which encodes the flow information for both the occluded and non-occluded points in the source. Finally, we use a similar predictor layer as in~\cite{wu2020pointpwc} to predict the residual flow for each source point, and the finer scene flow is the addition of the residual and the upsampled flow. The predicted scene flow and occlusion mask at pyramid level $l$ are further used in pyramid level ($l-1$) above it.

In this section, we mainly discuss the novel components in our model: Occlusion predictor and Occlusion-weighted Cost Volume. 
Implementation details and the schematic plots for all the components can be found in the supplementary.

\subsection{Occlusion Predictor}

Since the occluded regions usually produce misleading information for the flow estimation, they need special treatment in the architecture and training loss.
In our work, We use a small neural network to estimate the occlusion label for each point in the source. The occlusion predictor's inputs are the source point cloud, the warped target point cloud, and the upsampled occlusion label from the previous pyramid level. We use several 1$\times$1 convolutions to encode the similarity between each source point and its neighboring target points. Then we use a Max-pooling followed by MLP to generate the final occlusion label based on the encoded similarity between the point clouds. We also use a Sigmoid activation layer at the end to ensure the output $O(s_i)$ to be an occlusion probability with a value in the range [0,1] for each point in the source. 
Details and schematic plots of this layer can be found in the supplementary.


\begin{figure*}[t]
\begin{center}
\includegraphics[width=1.0\textwidth]{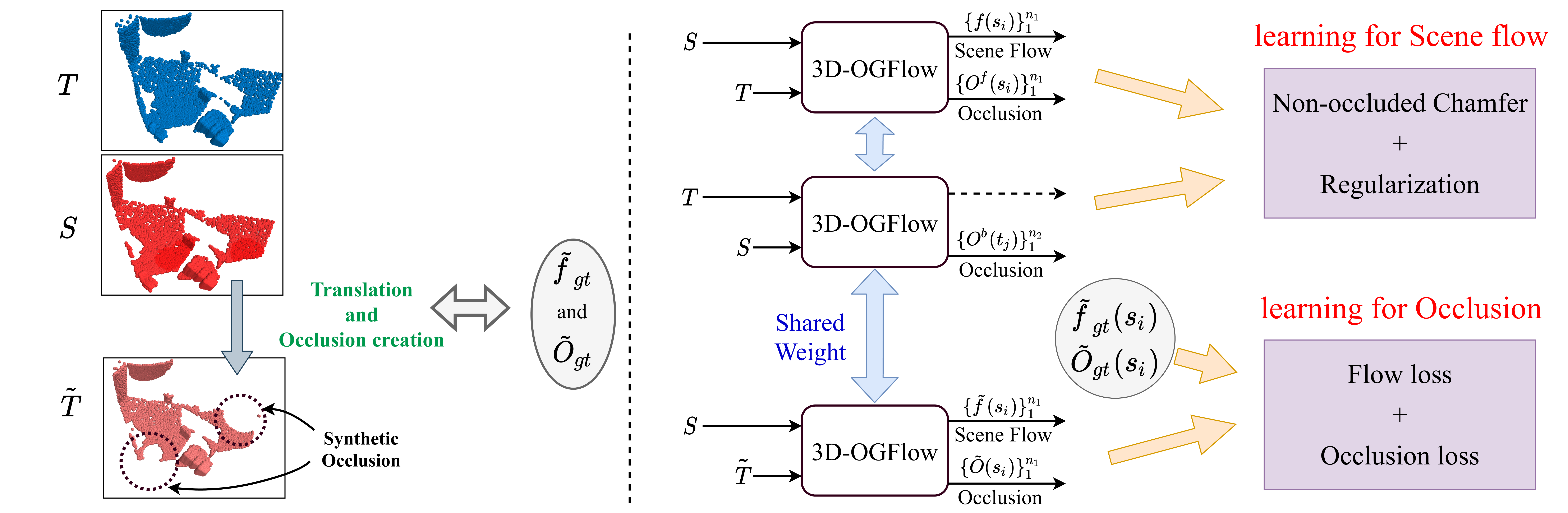}
\end{center}
\vspace{-15pt}
\caption{\textbf{Self-supervised Learning}. To learn the occlusion label, we create $\Tilde{T}$ from $S$ by applying a random translation $\Tilde{f}_{gt}$ and remove the k-NN of randomly selected points. The removed regions can be considered as occlusions and we obtain the occlusion label $\Tilde{O}_{gt}(s_i)$ for $s_i\in S$ with respect to $\Tilde{T}$. We make a forward and backward inference on ($S, T$) using our model to construct the non-occluded Chamfer loss and regularization. We make a third inference on ($S,\Tilde{T}$) and use $\Tilde{f}_{gt}$/$\Tilde{O}_{gt}$ as the ground truth supervision for the occlusion learning.}
\label{fig:self}
\end{figure*}

\subsection{Occlusion-weighted Cost Volume}
\label{sec:ppwoc_cv}

Cost Volume is a standard concept in stereo matching, it encodes the similarity and correlation between the consecutive time frames.~\cite{wu2020pointpwc} was the first to introduce Cost Volume's concept for the scene flow estimation on point clouds. However, their design does not consider the occlusion issues, and their model’s performance on the occluded scene can significantly decrease.~\cite{Ouyang_2021_CVPR} proposed an occlusion masking operation such that their Cost Volume for the occluded points becomes 0, but this can be harmful for the flow prediction for the occluded regions. 

In our Occlusion-weighted Cost Volume layer, we first construct the Matching Cost, it encodes the \textit{point-wise} correlation between a source point $s_i$ and a target point $t_j$. By using the source point feature $c_i$ and target point feature $g_j$, we calculate the Matching Cost between them by the following:

\begin{equation}
cost(s_i,t_j )=h(c_i,g_j,t_j-s_i)
\label{matchcost}
\end{equation}

\noindent
Where in $h(\cdot)$ we first concatenate all the inputs along the feature dimension, then we use several 1$\times$1 convolutions to process the data. 

After the Matching Cost construction, we construct the Cost Volume for $s_i$ by aggregating their Matching Cost with $t_j$ using Max-pooling. In order to avoid the expensive computation and high memory consumption, we only apply the aggregation among the K nearest neighbor (k-NN) in the warped target $T_w$ around $s_i$ ($N_{Tw}(s_i)$):

\begin{equation}
CV_{cross}(s_i)=\underset{t_j\in N_{Tw}(s_i)}{MAX}\{cost(s_i,t_j)\}
\label{cvcross}
\end{equation}

\noindent
For the occluded points in the source, which do not exist in the target frame, the CV construction in Eq.\ref{cvcross} based on the pair-wise cross-correlation might not be accurate. For such a point, its scene flow should be guided by its closest non-occluded points and consistent with its local nearest neighbors. By this motivation, we propose to construct a self Cost Volume by applying another self-aggregation:

\begin{equation}
CV_{self}(s_i)=\underset{s_k\in N_{S}(s_i)}{MAX}\{CV_{cross}(s_k)\}
\label{cvself}
\end{equation}

\noindent
Our final Cost Volume for each source point is the sum of the $CV_{cross}$ and $CV_{self}$ weighted by the predicted occlusion label:

\begin{equation}
CV(s_i)=O(s_i)CV_{cross}(s_i)+(1-O(s_i))CV_{self}(s_i)
\label{eq:cv}
\end{equation}

\noindent
where $O(s_i)$ is the predicted occlusion label for the $s_i$. The schematic plot can be found in the supplementary. 

Notice that during the \textit{supervised} training for the scene flow, to make the accurate flow prediction, the model needs to force the $CV_{cross}(\cdot)$ term in Eq.~\ref{eq:cv} to have a higher contribution for the non-occluded point $s_i$. While for the occluded point, the model needs to force the $CV_{self}(\cdot)$ term to have a higher contribution. Since the predicted occlusion $O(s_i)$ controls this weighting, it means our occlusion label is learned in a \textit{self-supervised} manner without explicit occlusion supervision during the \textit{supervised} training of the scene flow.

\section{Self-supervised Training}
Since the acquisition of the ground truth annotation is difficult or even impossible in many real-world scenarios, we present a self-supervised training method that does not require any ground truth scene flow or occlusion label. Most of the previous self-supervised methods~\cite{wu2020pointpwc,Kittenplon_2021_CVPR} use Chamfer distance loss with some regularization to move the source smoothly towards the target. Although these losses work perfectly on the non-occluded data, they often lead to an incorrect prediction of the flow when the scene contains occluded regions. This is because those methods do not exclude the occluded regions in the Chamfer distance calculation, so the occluded regions in one point cloud might map to the non-occluded regions in the other.
To discard the occluded points in the loss function, we need to know each source point's occlusion label, but this often requires an accurate scene flow prediction, which leads to a paradox. In our work, we suggest using another synthetic target point cloud to train the occlusion predictor.

For each source point cloud S in the dataset, we generate a synthetic target point cloud $\Tilde{T}$ from it by first applying a randomly generated translation to every point in the source. Then, we randomly choose several center points in the translated point cloud and remove their k-NN points from the translated point cloud, so that these removed regions can be considered as occluded. Since we generate the $\Tilde{T}$ from $S$ by ourselves, we know the ground truth scene flow $\Tilde{f}_{gt}(s_i)$ and the occlusion mask $\Tilde{O}_{gt}(s_i)$ from $S$ to $\Tilde{T}$. By training with this pair of point clouds ($S$, $\Tilde{T}$) using the supervised scene flow loss and occlusion loss, our model can learn to estimate the occlusion. Since we can never generate a real scene flow, the main goal of using ($S$, $\Tilde{T}$) is to learn the occlusion but not the scene flow. Due to the consideration of the expense in computation, we only use a simple rigid translation as our $\Tilde{f}_{gt}$ when constructing the $\Tilde{T}$ in our work. 

We also need to train our model with the regular pair of the point clouds ($S$, $T$) using the Non-occluded Chamfer distance with regularization to learn the scene flow. We formulate the overall self-supervised loss in Sec. \ref{sec:self}, and the general idea of this approach is shown in Fig.~\ref{fig:self}.

\begin{table*}[t!]
\centering
\small
\begin{tabular}{@{}c|l|c|ccccc@{}}
\toprule
Dataset                                              & \multicolumn{1}{c|}{Method} & Sup.          & EPE$_{full}{\downarrow}$ & EPE$\downarrow$ & ACC$_{05}{\uparrow}$ & ACC$_{10}{\uparrow}$ & Outliers$\downarrow$ \\ \midrule
\multicolumn{1}{l|}{\multirow{9}{*}{Flyingthings3D}} & FLOT(K=1)~\cite{puy20flot}                   & \textit{Full} & 0.2502                   & 0.1530           & 0.3965               & 0.6608               & 0.6625               \\
\multicolumn{1}{l|}{}                                & HPLFlowNet~\cite{gu_wang_wu_lee_wang_2019}                  & \textit{Full} & 0.2012                   & 0.1689          & 0.2629               & 0.5745               & 0.8123               \\
\multicolumn{1}{l|}{}                                & FlowNet3D~\cite{liu:2019:flownet3d}                   & \textit{Full} & 0.2119                   & 0.1577          & 0.2286               & 0.5821               & 0.8040                \\
\multicolumn{1}{l|}{}                                & OGSFNet~\cite{Ouyang_2021_CVPR}                     & \textit{Full} & 0.1634                   & 0.1217          & 0.5518               & 0.7767               & 0.5180                \\
\multicolumn{1}{l|}{}                                & PointPWC-Net~\cite{wu2020pointpwc}                & \textit{Full} & 0.1953                   & 0.1552          & 0.4160                & 0.6990                & 0.6389               \\
\multicolumn{1}{l|}{}                                & \textbf{Ours}               & \textit{Full} & \textbf{0.1383}          & \textbf{0.1031} & \textbf{0.6376}      & \textbf{0.8240}       & \textbf{0.4251}      \\ \cmidrule(l){2-8} 
\multicolumn{1}{l|}{}                                & ICP~\cite{10.1117/12.57955,chen_medioni}                         & \textit{Self} & 0.5048                   & 0.4848          & 0.1215               & 0.2558               & 0.9441               \\
\multicolumn{1}{l|}{}                                & PointPWC-Net~\cite{wu2020pointpwc}                & \textit{Self} & 0.6579                   & 0.3821          & 0.0489               & 0.1936               & 0.9741               \\
\multicolumn{1}{l|}{}                                & \textbf{Ours}               & \textit{Self} & \textbf{0.3373}          & \textbf{0.2796} & \textbf{0.1232}      & \textbf{0.3593}      & \textbf{0.9104}      \\ \midrule
\multirow{13}{*}{KITTI}                              & FLOT(K=1)~\cite{puy20flot}                   & \textit{Full} & 0.1303                   & -               & 0.2788               & 0.6672               & 0.5299               \\
                                                     & HPLFlowNet~\cite{gu_wang_wu_lee_wang_2019}                  & \textit{Full} & 0.3430                    & -               & 0.1035               & 0.3867               & 0.8142               \\
                                                     & FlowNet3D~\cite{liu:2019:flownet3d}                   & \textit{Full} & 0.1834                   & -               & 0.0980                & 0.3945               & 0.7993               \\
                                                     & OGSFNet~\cite{Ouyang_2021_CVPR}                     & \textit{Full} & 0.0751                   & -               & 0.7060                & 0.8693               & 0.3277               \\
                                                     & PointPWC-Net~\cite{wu2020pointpwc}                & \textit{Full} & 0.1180                    & -               & 0.4031               & 0.7573               & 0.4966               \\
                                                     & \textbf{Ours}               & \textit{Full} & \textbf{0.0595}          & -               & \textbf{0.7755}      & \textbf{0.9069}      & \textbf{0.2732}      \\ \cmidrule(l){2-8} 
                                                     & ICP~\cite{10.1117/12.57955,chen_medioni}                         & \textit{Self} & 0.3801                   & -               & 0.1038               & 0.2913               & 0.8307               \\
                                                     & PointPWC-Net~\cite{wu2020pointpwc}                & \textit{Self} & 0.3373                   & -               & 0.0529               & 0.2125               & 0.9352               \\
                                                     & \textbf{Ours}               & \textit{Self} & \textbf{0.2091}          & -               & \textbf{0.2107}      & \textbf{0.4904}      & \textbf{0.7241}      \\ \cmidrule(l){2-8} 
                                                     & PointPWC-Net~\cite{wu2020pointpwc}                & \textit{Full}+\textit{Full ft}  & 0.0650                    & -               & 0.6618               & 0.8894               & 0.3569               \\
                                                     & \textbf{Ours}               & \textit{Full}+\textit{Full ft}  & \textbf{0.0249}          & -               & \textbf{0.9335}      & \textbf{0.9721}      & \textbf{0.1498}      \\ \cmidrule(l){2-8} 
                                                     & PointPWC-Net~\cite{wu2020pointpwc}                & \textit{Self}+\textit{Self ft}  & 0.1632                   & -               & 0.2117               & 0.5409               & 0.6934               \\
                                                     & \textbf{Ours}               & \textit{Self}+\textit{Self ft}  & \textbf{0.0857}          & -               & \textbf{0.5146}      & \textbf{0.8108}      & \textbf{0.4724}      \\ \bottomrule
\end{tabular}
\vspace{-5pt}
\caption{\textbf{Evaluation on occluded Flyingthings3D and KITTI}. For Flyingthings3D, \textit{Full} / \textit{Self} means the training is done using supervised/self-supervised schemes. For KITTI, \textit{Full} / \textit{Self} means we evaluate the corresponding trained models from Flyingthings3D directly on the KITTI without fine-tuning. We also perform 2 kinds of fine-tuning experiments on the bottom: \textit{Full}+\textit{Full ft} means we apply the \textit{supervised} fine-tuning on the \textit{supervised} pretrained weight on Flyingthings3D, and similarly for the \textit{Self}+\textit{Self ft}. Our model outperforms the previous methods by a large margin on all kinds of supervision and evaluation metrics.}
\label{tab:ppwoc_f3d_kitti}
\end{table*}

\section{Loss functions}

\subsection{Self-supervised Loss}
\label{sec:self}

\noindent
\textbf{Flow/Occlusion loss (synthetic)}. To train the occlusion predictor in a self-supervised manner, we create a synthetic target $\Tilde{T}$ as explained in the previous section. Since the occlusion prediction at each pyramid level depends on the upsampled scene flow from its previous level, we also need the flow loss in addition to the occlusion loss. By using the synthetic ground truth scene flow $\Tilde{f}_{gt}$ and occlusion mask $\Tilde{O}_{gt}$ from $S$ to $\Tilde{T}$, we construct the following multi-level “supervised” losses:

\begin{equation}
\Tilde{L}_f(S,\Tilde{T},\Theta)= \sum\limits_{l=0}^{3}\alpha_l\sum\limits_{s_i\in S^l}\|\Tilde{f}_{gt}(s_i)-\Tilde{f}(s_i)\|_2
\label{flowloss}
\end{equation}

\begin{equation}
\Tilde{L}_{oc}(S,\Tilde{T},\Theta)= \sum\limits_{l=0}^{3}\alpha_l\sum\limits_{s_i\in S^l}\|\Tilde{O}_{gt}(s_i)-\Tilde{O}(s_i)\|
\label{occloss}
\end{equation}

\noindent
Where $S$ and $\Tilde{T}$ are the inputs to the model, $S^l$ is the downsampled source point cloud at pyramid level $l$, $\Theta$ is all the learnable parameters of the model, and $\Tilde{f}(s_i)$ and $\Tilde{O}(s_i)$ are the predicted flow and occlusion from $s_i$ to $\Tilde{T}$.

\begin{figure*}
\begin{center}
\includegraphics[width=1.0\textwidth]{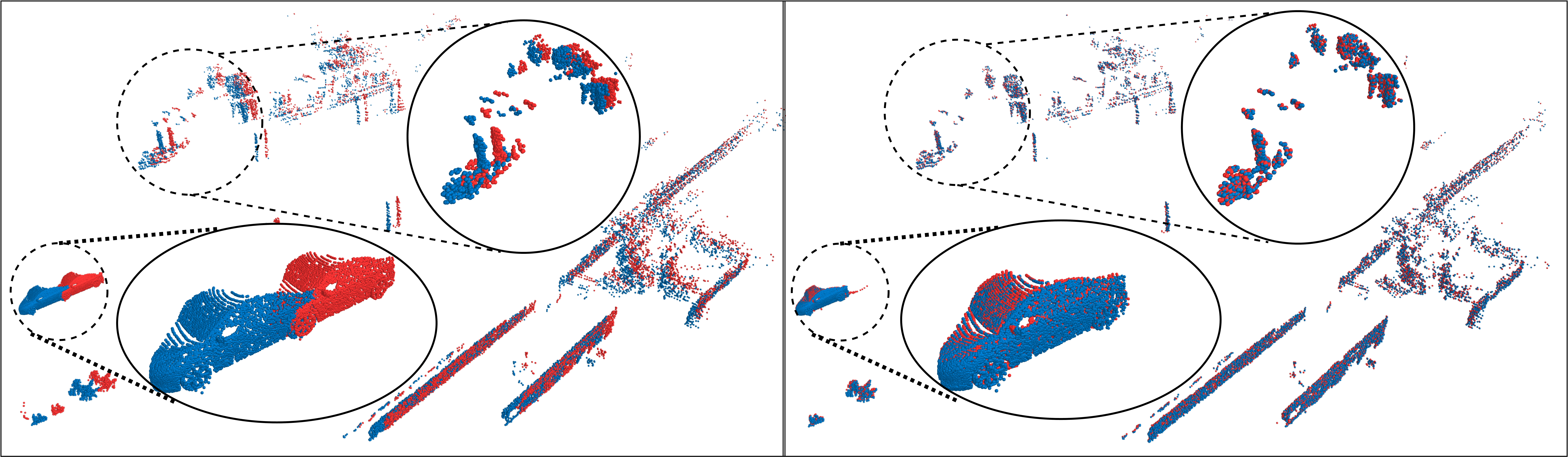}
\end{center}
    \vspace{-13pt}
   \caption{\textbf{Visualization on KITTI.} We plot the source (red) and the target (blue) frames from the KITTI on the same 3D space on the left. On the right, we show the warped source point cloud according to the estimated flow (source+flow) learned from the supervised training and the target point cloud. We also provide the zoomed view for the circled region to show the alignment better.} 
\label{fig:ockitti}
\end{figure*}

\noindent
\textbf{Non-occluded Chamfer Distance.} As explained in the previous section, it is crucial to exclude the occluded region in the chamfer distance calculation. In order to minimize the distances between the non-occluded source points and the non-occluded target points, we use the following loss:
\begin{align*}
\begin{split}
D_l(S^l,T^l,\Theta)=|S^l|\sum\limits_{s_i\in S^l_w}\frac{\underset{t_j\in T^l}{min}\|s_i-t_j\|_2\cdot O^{f}(s_i)}{\sum\limits_{s_i\in S^l_w}O^{f}(s_i)} \\
                     +|T^l|\sum\limits_{t_j\in T^l}\frac{\underset{s_i\in S^l_w}{min}\|t_j-s_i\|_2\cdot O^{b}(t_j)}{\sum\limits_{t_j\in T}O^{b}(t_j)}
\end{split}
\end{align*}
\vspace{-10pt}
\begin{equation}
L_{nch}(S,T,\Theta)=\sum\limits_{l=0}^{3}\alpha_l D_l(S^l,T^l,\Theta)
\label{chamfer}
\end{equation}

\noindent
Where $O^{f}$ and $O^{b}$ are the predicted forward ($S$$\rightarrow$$T$) and backward ($T$$\rightarrow$$S$) occlusion label, $|S^l|$ and $|T^l|$ are the numbers of points in $S^l$ and $T^l$, $S^l_w$=$\{s_i+f(s_i)|s_i\in S^l\}_{i=1}^{|S^l|}$ is the warped source point cloud at each pyramid level $l$ according to the predicted flow. The backward occlusion $O^b(t_j)$ from the target towards the source is obtained by simply swapping the order of inputs ($S, T$). In order to avoid a degenerated solution where \textbf{all} the points are predicted as occluded ($O^f(s_i)=O^b(t_j)=0$), we remove $O^{f}(s_i)$ and $O^{b}(t_j)$ from the computational graph during the backpropagation with SGD. In other words, when we update the model's parameters according to the gradient of non-occluded Chamfer distance, we exclude the parameters of the occlusion predictor and only update the rest of it.

\noindent
\textbf{Smoothness regularization.} In addition to the non-occluded Chamfer distance, we also need a regularization to ensure that the predicted scene flow is smooth among the local neighbourhoods. Since the flow of the occluded points should be consistent with and guided by the flow of its nearest neighbors, we do not exclude the occluded regions in this smoothness regularization. Following ~\cite{wu2020pointpwc}, our regularization term is defined as:
\begin{multline*}
R^{l}(S^l,T^l,\Theta)= \sum\limits_{s_i\in S^l}\sum\limits_{s_k\in N_S(s_i)}\frac{\|f(s_i)-f(s_k)\|_1}{|N_S(s_i)|}
\end{multline*}
\vspace{-10pt}
\begin{equation}
L_{reg}(S,T,\Theta)=\sum\limits_{l=0}^{3}\alpha_l R^{l}(S^l,T^l,\Theta)    
\end{equation}

\noindent
Where $|N_S(s_i)|$ is the number of points in the neighbourhood $N_S(s_i)$.

The overall self-supervised loss is the weighted sum of these four losses with the scale factor $\lambda_{reg},\lambda_{f},\lambda_{oc}$:
\begin{equation}
L_{self}=L_{nch}+\lambda_{reg}L_{reg}+\lambda_f\Tilde{L}_f+\lambda_{oc}\Tilde{L}_{oc}
\label{eq:self}
\end{equation}

\subsection{Fully-supervised Loss}
\label{sec:super}
We use similar multi-level flow loss in Eq.\ref{flowloss} with the ground truth scene flow for the supervised training. Due to our occlusion-weighted design in Sec~\ref{sec:ppwoc_cv}, the model does not require explicit occlusion loss for the supervised learning since it can extract the occlusion information from the ground truth scene flow.
Our supervised scene flow loss is formulated below:
\begin{equation}
L_{sup}(S,T,\Theta)= \sum\limits_{l=0}^{3}\alpha_l\sum\limits_{s_i\in S^l}\|f_{gt}(s_i)-f(s_i)\|_2
\label{eq:super}
\end{equation}
\noindent
where $f_{gt}(s_i)$ and $f(s_i)$ are the ground truth and predicted scene flow from $s_i\in S$ to $T$.

\section{Experiments}
By following the experimental procedure as in~\cite{liu:2019:flownet3d,puy20flot,wu2020pointpwc}, we first train our model on the \textit{occluded} Flyingthings3D~\cite{mayer_ilg_hausser_fischer_cremers_dosovitskiy_brox_2016} dataset using the supervised and self-supervised training schemes (Sec. \ref{sec:ppwoc_f3d}). Then, we evaluate the trained models from the two training schemes on the real LiDAR scans from the \textit{occluded} KITTI scene flow 2015~\cite{menze_geiger_2015,menze_heipke_geiger_2015} with and without fine-tuning (Sec. \ref{sec:ppwoc_kitti}). This kind of evaluation procedure is a common practice in the previous works since it is difficult to acquire scene flow from real data and the KITTI dataset is too small for the training. Notice that the two datasets we are using are pre-processed by~\cite{liu:2019:flownet3d}, where the scene in both datasets contains \textbf{occlusion} up to some degree. The Flyingthings3D datasets contain 20000 pairs of point clouds in the training set and 2000 in the validation set. Each pair in the dataset contains two point clouds representing the sampled 3D synthetic scene at two different time frames, and the scene in this dataset is \textit{highly} occluded. The KITTI dataset contains the LiDAR scans of the real scene with some occlusions, and it contains 150 pairs of point clouds. We provide more visualization results and explanations on the two datasets in the supplementary. In Sec. \ref{sec:ppwoc_abl}, we present several ablation experiments to validate our novel design in the model and our self-supervised losses. Finally, in Sec. \ref{sec:ppwoc_noc}, to show our model's expressiveness, we compare our model with the previous work on the \textit{non-occluded} datasets.

\noindent
\textbf{Implementation details.} Our model uses the Feature Pyramid Network and Upsample layer proposed by~\cite{wu2020pointpwc} and uses the Warping layer as in~\cite{Ouyang_2021_CVPR}. Implementation details of our architecture can be found in the supplementary. We use $n_1=n_2=8192$ points for each point cloud in the two datasets when performing all kinds of experiments. The weights $\{\alpha_0,...,\alpha_3\}$ in the loss functions are $\{0.02,0.04,0.08,0.16\}$. For the \textbf{supervised} training on Flyingthings3D, we use 2 GTX2080Ti GPU with a batch size of 8. We train it for 120 epochs with an initial learning rate of 0.001, and we reduce it with a decay rate of 0.85 after every 10 epochs. We further reduce the decay rate from 0.85 to 0.8 after 75 epochs. For the \textbf{self-supervised} training on Flyingthings3D, we use 8 GTX2080Ti GPU with a batch size of 24. In Eq.~\ref{eq:self}, we choose $\lambda_f=0.6$ and $\lambda_{oc}=1.0$. We set $\lambda_{reg}=3.0$ in the first 50 epochs, and then we reduce it to 1.0 gradually from 50 to 70 epochs. We train the model for 150 epochs with an initial learning rate of 0.001 and a decay rate of 0.83 for every 10 epochs. In the first 30 epochs, we use Eq.~\ref{eq:self} as the loss function. After 30 epochs, we exclude the synthetic flow loss and only use the rest of the 3 terms in Eq.~\ref{eq:self} for the self-supervised training. The magnitude of the randomly generated translation ($\Tilde{f}_{gt}(s_i)$) is 2 meters.

\noindent
\textbf{Metrics.} We follow the same evaluation metrics as in ~\cite{puy20flot,liu:2019:flownet3d,Ouyang_2021_CVPR} to evaluate the performance of the different models:

\begin{itemize}
\item $EPE_{full}$(m): $\big\|f_{gt}(s_i)-f(s_i)\big\|_2$ averaged over \textbf{all} $p_i\in S$.
\item $EPE$(m): $\big\|f_{gt}(s_i)-f(s_i)\big\|_2$ averaged over all \textbf{non occluded} $p_i\in S$.
\item $ACC_{05}$: percentage of points whose $EPE$ $<$ 0.05m or relative error$<$ 5\%
\item $ACC_{10}$: percentage of points whose $EPE$ $<$ 0.1m or relative error$<$ 10\%
\item $Outlier$: percentage of points whose $EPE$ $>$ 0.3m or relative error$>$ 10\%

\end{itemize}

\begin{figure}
\begin{center}
\includegraphics[width=0.45\textwidth,keepaspectratio]{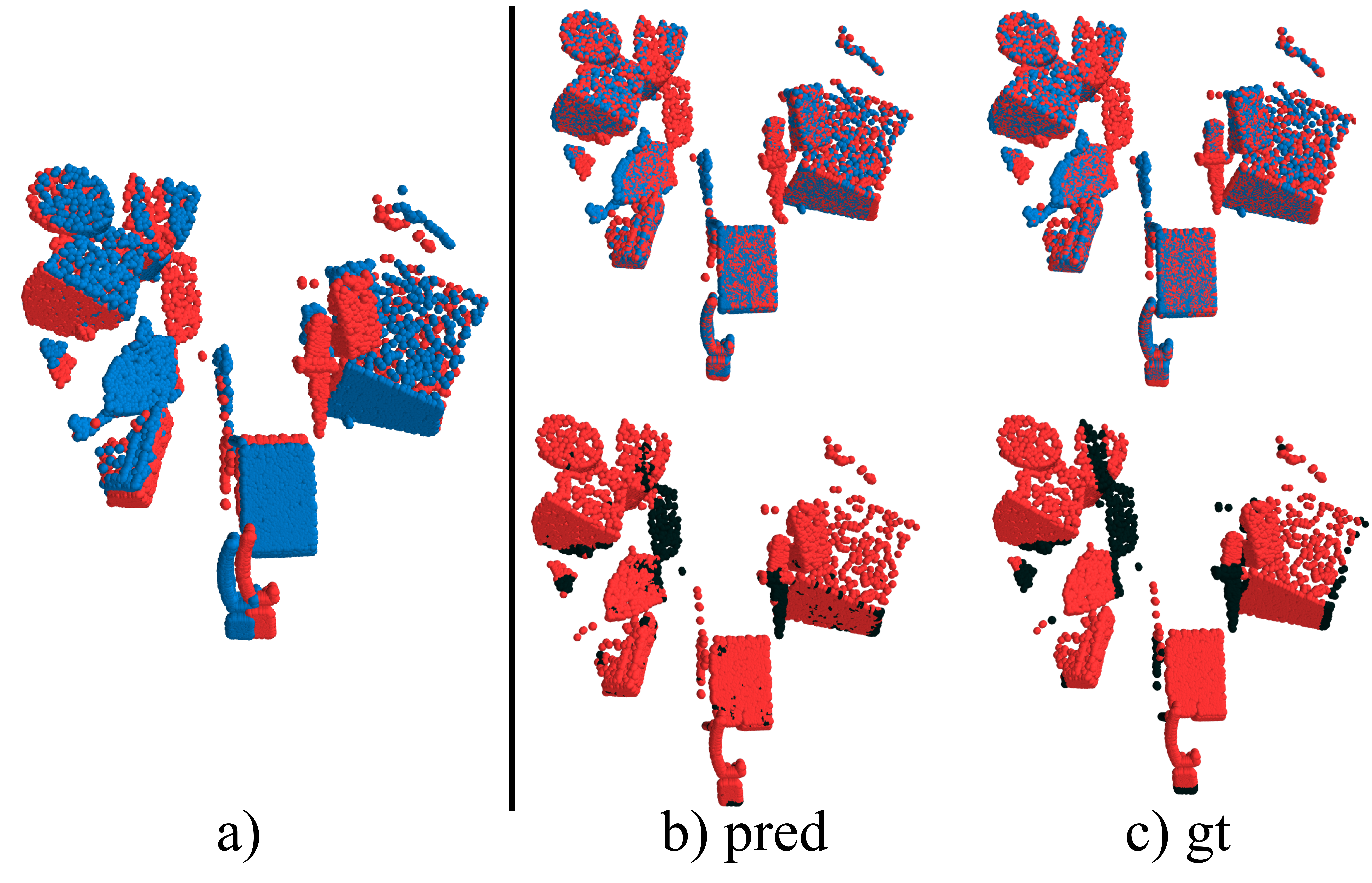}
\end{center}
    \vspace{-14pt}
   \caption{\textbf{Visualization on Flyingthings3D.} In a), we plot the source (red) and the target (blue) of a test sample on the same 3D space. In b) and c), we plot the warped source according to predicted/ground truth flow (source+flow) and the target on the top, we can see that warped source aligned to the target. On the bottom, we show the predicted/ground truth occlusion map, where black points mean occluded and red points mean non-occluded.} 
\label{fig:ocf3d}
\end{figure}

\subsection{Evaluation on Flyingthings3D}
\label{sec:ppwoc_f3d}
We first train our model using the supervised (Sec. \ref{sec:super}) and self-supervised (Sec. \ref{sec:self}) frameworks on the training set of Flyingthings3D, then we test the corresponding models on the validation set. The results are reported in Table~\ref{tab:ppwoc_f3d_kitti}. We can see that our method has the best performance on all metrics. For fully-supervised training, we outperform~\cite{Ouyang_2021_CVPR} by 15.3\% on EPE$_{full}$. Our self-supervised system outperforms~\cite{wu2020pointpwc} by 48\% on EPE$_{full}$, which clearly shows the power of our self-supervised training schemes. 
We want to emphasize that the reported numbers for the  PointPWC-Net~\cite{wu2020pointpwc} and HPLFlowNet~\cite{gu_wang_wu_lee_wang_2019} in Table~\ref{tab:ppwoc_f3d_kitti} are \textbf{different} from the one reported by their own paper, this is because we evaluate all the models in Table~\ref{tab:ppwoc_f3d_kitti} on the \textit{occluded} Flyingthings3D and \textit{occluded} KITTI proposed by~\cite{liu:2019:flownet3d}, while~\cite{wu2020pointpwc,gu_wang_wu_lee_wang_2019} only evaluate on \textit{non-occluded} version proposed by~\cite{gu_wang_wu_lee_wang_2019}. The evaluation results on the non-occluded version of datasets can be found in Table. \ref{tab:noc}


For the occlusion estimation, our model achieves a 92.3\% accuracy on the validation set of Flyingthings3D under the fully-supervised training scheme. When we train the model using our self-supervised losses, where the occlusion predictor is purely trained from the synthetic occlusion, we can still achieve a 90.9\% accuracy. These results demonstrate the generalization ability of our model to unseen data with real occlusion under both the supervised and self-supervised frameworks. Visualizations are shown in Fig.~\ref{fig:ocf3d}.

\begin{table*}
\hspace{-20pt}
\small
\begin{subtable}{.5\linewidth}
\centering
\begin{tabular}{@{}l|c@{}}
\toprule
\multicolumn{1}{c|}{Cost Volume} & EPE$\downarrow$ \\ \midrule
$CV_{cross}(\cdot)$              & 0.1352          \\
$CV_{self}(\cdot)$               & 0.1324          \\
$O(\cdot)CV_{cross}(\cdot)$      & 0.1242          \\
Occ weighted                     & \textbf{0.1031} \\ \bottomrule
\end{tabular}
\caption{Model Design}
\end{subtable}%
\begin{subtable}{.5\linewidth}
\centering
\begin{tabular}{@{}c|ccc|c@{}}
\toprule
Chamfer      & Reg. & Occ ($Syn.$) & Flow ($Syn.$) & EPE$_{full}{\downarrow}$ \\ \midrule
Regular      & \XSolidBrush   & \XSolidBrush            & \XSolidBrush             & 0.9071                   \\
Regular      & \Checkmark   & \XSolidBrush            & \XSolidBrush             & 0.4696                   \\
Non-occluded & \Checkmark   & \Checkmark            & \XSolidBrush             & 0.4085                   \\
Non-occluded & \Checkmark   & \Checkmark            & \Checkmark             & \textbf{0.3373}          \\ \bottomrule
\end{tabular}
\vspace{-2pt}
\caption{Self-supervised losses}
\end{subtable}
\vspace{-5pt}
\caption{\textbf{Ablation Study.} In (a), we test different design choices of the Cost Volume, and our occlusion-weighted design gives the best performance. In (b), we compare the regular and non-occluded Chamfer loss and show each term's usefulness in our self-supervised losses.}
\label{tab:ppwoc_abl}
\end{table*}

\subsection{Evaluation on KITTI}
\label{sec:ppwoc_kitti}

In this section, we first evaluate the supervised and self-supervised pretrained models from Flyingthings3D directly on all the 150 samples from KITTI without fine-tuning. Then, we perform the fine-tuning experiments on KITTI by using the pretrained weight from Flyingthings3D. We split the KITTI into 100 samples of the training set for the fine-tuning, 50 samples of the test set. The numbers are shown in Table~\ref{tab:ppwoc_f3d_kitti}. Since the KITTI does not provide the ground truth occlusion mask, we cannot evaluate the EPE on this dataset. Some visualizations are shown in Fig.\ref{fig:ockitti}.

\noindent
\textbf{Generalization results.} The direct evaluation results are shown in the upper part of the KITTI section in Table~\ref{tab:ppwoc_f3d_kitti}. For both the supervised and self-supervised frameworks, our model has the best generalization ability.

\noindent
\textbf{Fine-tuned results.} The fine-tuning results are shown in the lower part of the KITTI section in Table~\ref{tab:ppwoc_f3d_kitti}. We first perform the supervised fine-tuning (\textit{Full ft}) using Eq.~\ref{eq:super} on the training set of KITTI using the supervised (\textit{Full}) pretrained weight on Flyingthings3D. Then, we perform the self-supervised fine-tuning (\textit{Self ft}) using Eq.~\ref{eq:self} on the training set of KITTI using the self-supervised (\textit{Self}) pretrained weight on Flyingthings3D without using any kinds of ground truth. We compare our results with~\cite{wu2020pointpwc}, and we can see that our supervised/self-supervised fine-tuning can give much larger improvements in the performance.



\subsection{Ablation Study}
\label{sec:ppwoc_abl}

In Table~\ref{tab:ppwoc_abl} (a), we perform several ablation experiments to validate the novel design of our Cost Volume layer. We train the model with the corresponding design on the Flyingthings3D, and we report their EPE on the validation set. When we use the cross-correlation term $CV_{cross}$ as our Cost Volume, we achieve a 0.1352 EPE. If we exclude the occluded regions in the $CV_{cross}$ by masking with the predicted occlusion ($O(s_i)\cdot CV_{cross}(s_i)$) as in~\cite{Ouyang_2021_CVPR}, we achieve a 8\% improvement in EPE. In the last row, our occlusion weighted design in Eq.\ref{eq:cv} gives the best results.

In Table~\ref{tab:ppwoc_abl} (b), we test the performance of our model trained by different combinations of self-supervised losses and present the EPE$_{full}$ on the validation set of Flyingthings3D. Using the regular Chamfer loss and smoothness regularization, we get an EPE$_{full}$ of 0.4696. When we exclude the occluded regions in the Chamfer loss by using the predicted occlusion label learned from synthetic occlusion loss, the performance improves. When we add the synthetic flow loss, our occlusion learning can be more accurate, and so is the occlusion elimination in the non-occluded Chamfer loss. This design gives the best performance.

\begin{table}
\footnotesize
\centering
\setlength\tabcolsep{4pt}
\begin{tabular}{@{}l|l|ccc@{}}
\toprule
\multicolumn{1}{c|}{Datasets}  & \multicolumn{1}{c|}{Method} & EPE$_{full}{\downarrow}$ & ACC$_{05}{\uparrow}$ & \multicolumn{1}{l}{Outliers$\downarrow$} \\ \midrule
\multirow{4}{*}{Flyingthing3D} & FlowNet3D~\cite{liu:2019:flownet3d}                   & 0.1136                   & 0.4125               & 0.6016                                   \\
                               & HPLFlowNet~\cite{gu_wang_wu_lee_wang_2019}                  & 0.0804                   & 0.6144               & 0.4287                                   \\
                               & PointPWC-Net~\cite{wu2020pointpwc}                & 0.0588                   & 0.7379               & 0.3424                                   \\
                               & \textbf{Ours}               & \textbf{0.0360}          & \textbf{0.8790}      & \textbf{0.1969}                          \\ \midrule
\multirow{4}{*}{KITTI}         & FlowNet3D~\cite{liu:2019:flownet3d}                   & 0.1767                   & 0.3738               & 0.5271                                   \\
                               & HPLFlowNet~\cite{gu_wang_wu_lee_wang_2019}                  & 0.1169                   & 0.4783               & 0.4103                                   \\
                               & PointPWC-Net~\cite{wu2020pointpwc}                & 0.0694                   & 0.7281               & 0.2648                                   \\
                               & \textbf{Ours}               & \textbf{0.0385}          & \textbf{0.8817}      & \textbf{0.1754}                          \\ \bottomrule
\end{tabular}
\caption{\textbf{Evaluation on non-occluded datasets.} We evaluate our methods on the non-occluded version of Flyingthings3D and KITTI used by~\cite{gu_wang_wu_lee_wang_2019,wu2020pointpwc}. Both models are trained using the supervised loss.}
\label{tab:noc}
\end{table}

\subsection{Evaluation on non-occluded datasets}
\label{sec:ppwoc_noc}

Many of the previous works~\cite{gu_wang_wu_lee_wang_2019,wu2020pointpwc} evaluated their methods only on the \textit{non-occluded} version of Flyingthings3D and KITTI proposed by~\cite{gu_wang_wu_lee_wang_2019}, where the occluded regions in the point clouds are removed during pre-processing, accordingly, we also evaluate 3D-OGFlow on these \textit{non-occluded} datasets to show the robustness of our model. By following the same procedure as in~\cite{gu_wang_wu_lee_wang_2019,wu2020pointpwc}, we train all the models
on the non-occluded Flyingthings3D in a fully-supervised manner and then directly evaluate them on non-occluded KITTI without any fine-tuning. The results are shown in Table~\ref{tab:noc}, and we can see that we are also winning on these non-occluded datasets.

\section{Conclusion}
In this work, we present a neural network with a novel occlusion-aware correlation layer for the scene flow estimation on the point clouds. Our occlusion weighted Cost Volume layer can compute the correlation between the point clouds for both the occluded and non-occluded regions according to their properties. We conduct several ablations to validate our occlusion weighted design of the Cost Volume. As a side benefit, by only using the ground truth scene flow as the supervision, this design can also let us learn the occlusion in a self-supervised manner. We also propose a novel self-supervised training scheme to learn the scene flow for the occluded scene efficiently. We achieve state-of-the-art performance on multiple datasets for both the supervised and self-supervised training schemes.

{\small
\bibliographystyle{ieee_fullname}
\bibliography{main}
}

\section{Supplementary}

\subsection{Details of the Architecture}
\begin{figure}
\begin{center}
\includegraphics[width=0.5\textwidth]{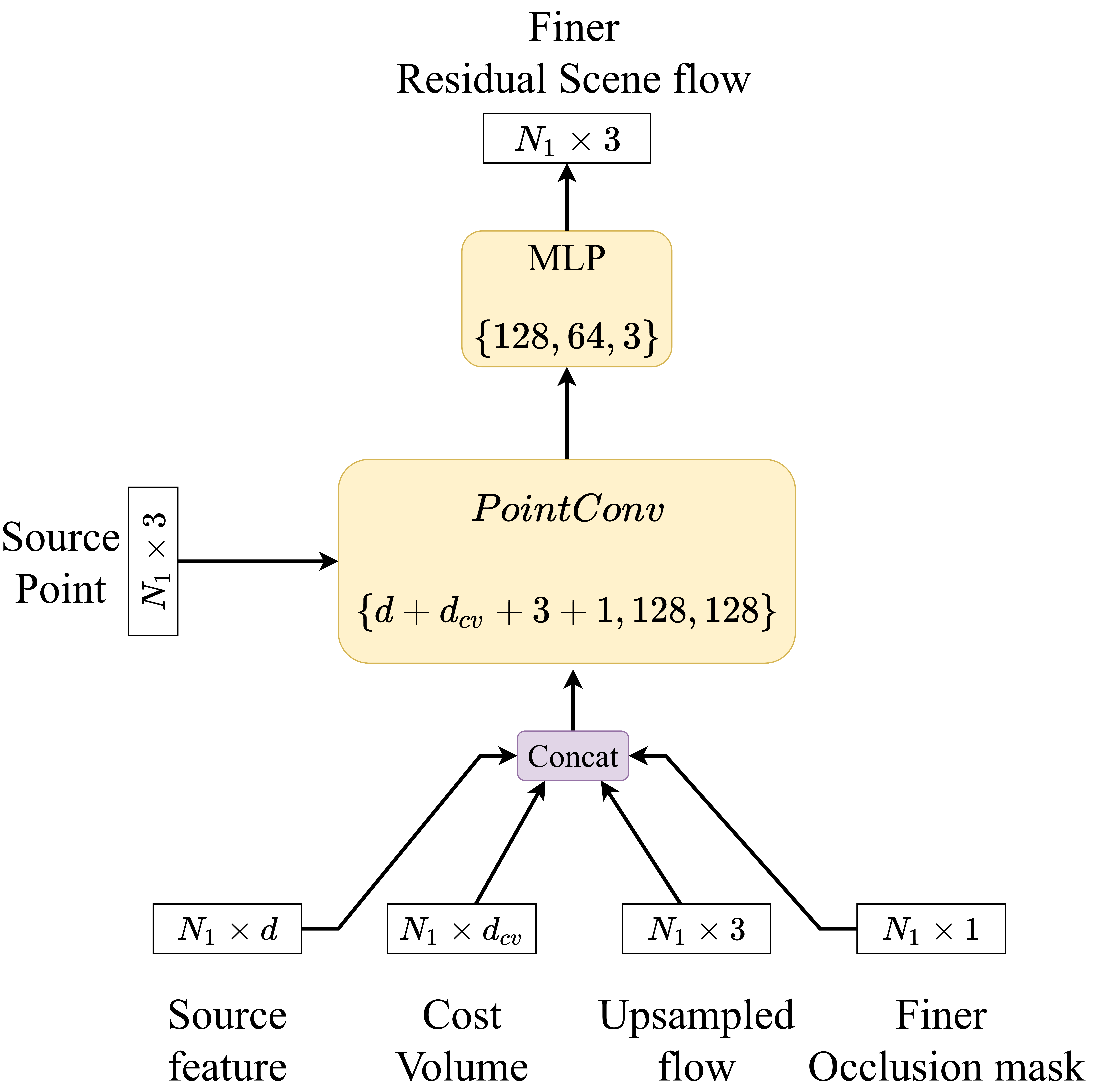}
\end{center}
\vspace{-15pt}
\caption{\textbf{Residual flow predictor}. We first concatenate the source feature, cost volume, upsampled flow from the previous pyramid level, and predicted occlusion mask at the current level along the feature dimension. Then, we use the PointConv to perform the feature encoding. Finally, we use MLP to produce the finer residual scene flow.}
\label{fig:ppwoc_flow}
\end{figure}

\begin{figure*}
\begin{center}
\includegraphics[width=1.0\textwidth]{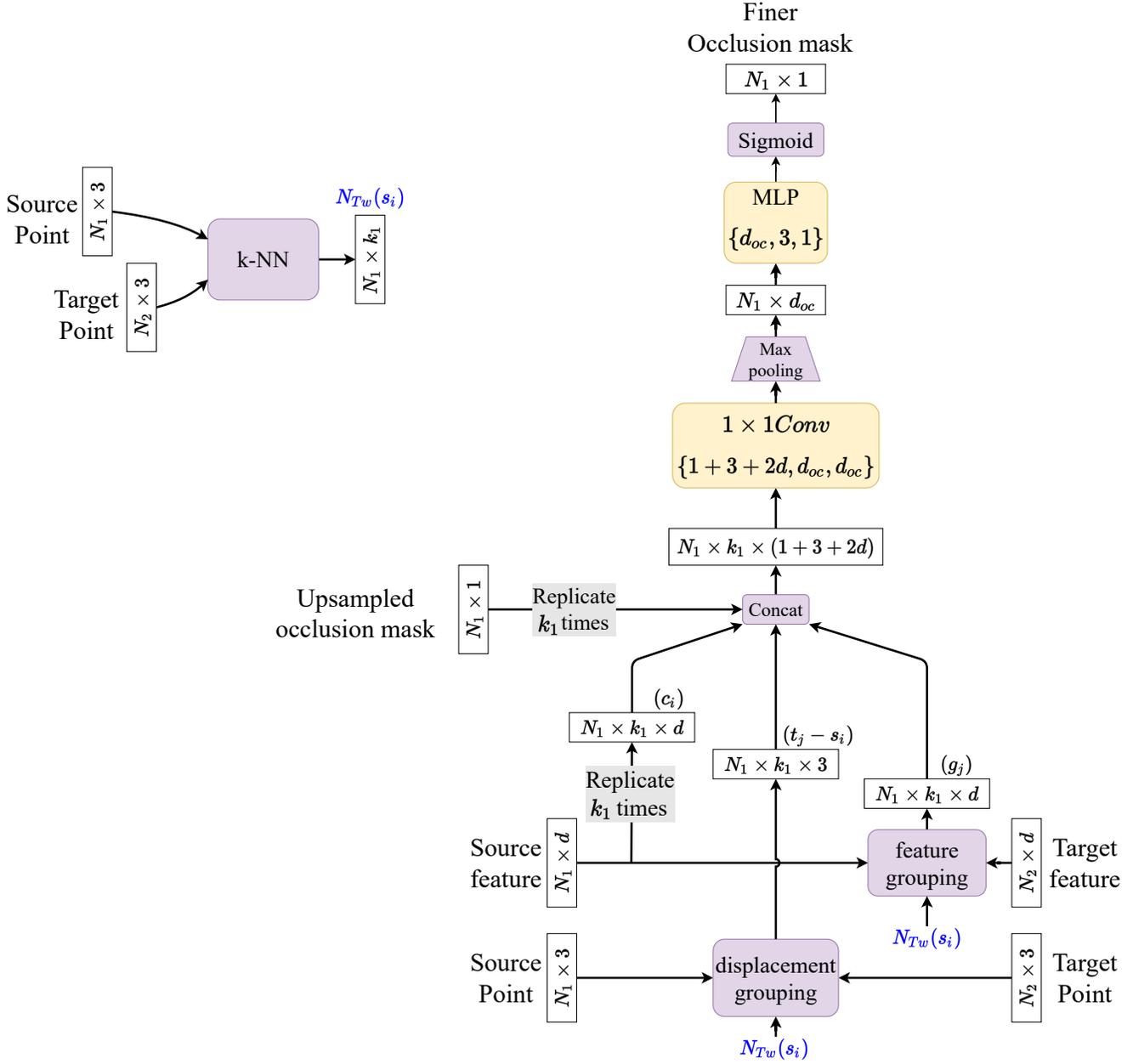}
\end{center}
\vspace{-20pt}
\caption{\textbf{Occlusion predictor}. The modules in purple represent the operation without any learnable parameters. The modules in yellow represent the operation with learnable parameters. The shape of the intermediate tensors is also provided. We first find the k-NN index ($N_{Tw}(s_i)$) in the warped target $T_w$ for each $s_i\in S$. Second, we gather the relative displacement ($t_j-s_i$) and the target features ($g_j$) using the k-NN index. We concatenate the upsampled occlusion mask, source features $c_i$, relative displacement $t_j-s_i$, and target features $g_j$ along the feature dimension. By using the 1$\times$1 convolutions, Max-pooling, and multilayer perceptron (MLP), we obtain the predicted finer occlusion mask. We also connect a Sigmoid activation layer at the end to ensure the output has values within the range [0,1].}
\label{fig:ppwoc_occ}
\end{figure*}

\begin{figure*}
\begin{center}
\includegraphics[width=1.0\textwidth]{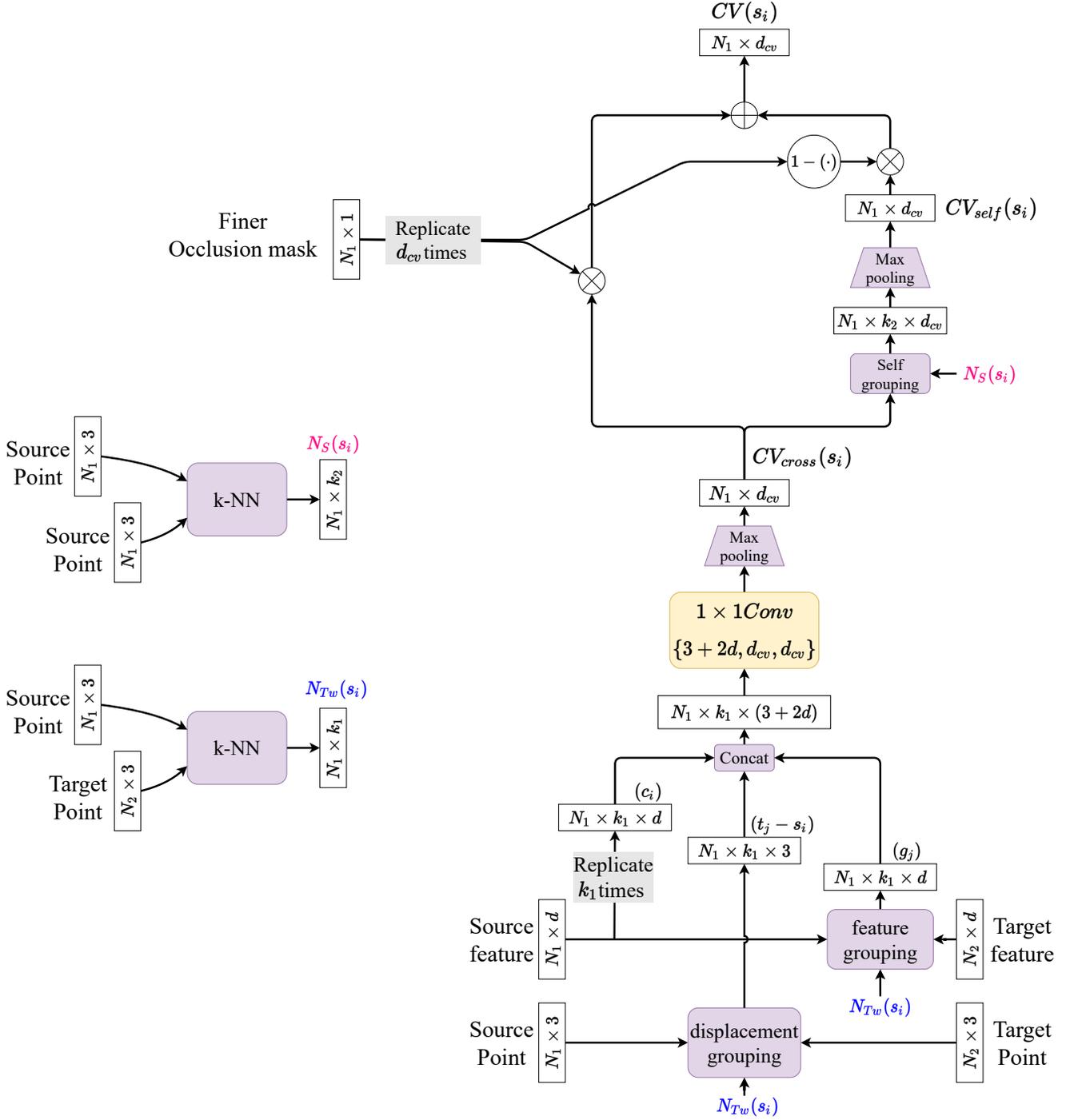}
\end{center}
\vspace{-20pt}
\caption{\textbf{Occlusion-weighted cost volume layer}. The modules in purple represent the operation without any learnable parameters. The modules in yellow represent the operation with learnable parameters. The shape of the intermediate tensors is also provided. We first find the k-NN index ($N_{Tw}(s_i)$) in the warped target $T_w$ for each $s_i\in S$. Then, we find the nearest local neighbors ($N_S(s_i)$) in the source for each $s_i$. We gather the relative displacement ($t_j-s_i$) and the target features ($g_j$) using the k-NN index $N_{Tw}(s_i)$, and then we concatenate these intermediate tensors along the feature dimension. We construct the matching cost between ($s_i$,$t_j$) using the 1$\times$1 convolutions. By using the Max-pooling as aggregation function, we obtain $CV_{cross}$. By further applying a self-aggregation on the $CV_{cross}$, we obtained the $CV_{self}$. The final cost volume is simply the weighted sum of the two terms using the predicted occlusion as explained in the paper.}
\label{fig:ppwoc_cv}
\end{figure*}

The details of the Occlusion predictor are shown in the figure~\ref{fig:ppwoc_occ}. The details of the Occlusion-weighted cost volume are shown in the figure~\ref{fig:ppwoc_cv}. We set $d_{cv}$ in the cost volume layer to be [32, 64, 128, 256] at pyramid level $l=0,1,2,3$. We set $d_{oc}$ in the occlusion predictor to be 64 at all pyramid levels. The numbers of the nearest neighbors we are using are $k_1=32$ and $k_2=64$. Notice that the beginning part (relative displacement/feature grouping) of these two layers have the same structure, so they are being shared in our implementation to reduce the running time. The details of the Residual flow predictor are shown in the figure~\ref{fig:ppwoc_flow}. The predicted finer flow at each pyramid level $l$ is simply the element-wise addition of the residual and the upsampled flow from level $l+1$. In our implementation, we set the length of the point cloud features $d$ to be [64, 96, 192, 320] at pyramid level $l=0,1,2,3$.

\subsection{More Visualization}
\begin{figure*}
\begin{center}
\vspace{-25pt}
\includegraphics[width=1.0\textwidth,keepaspectratio]{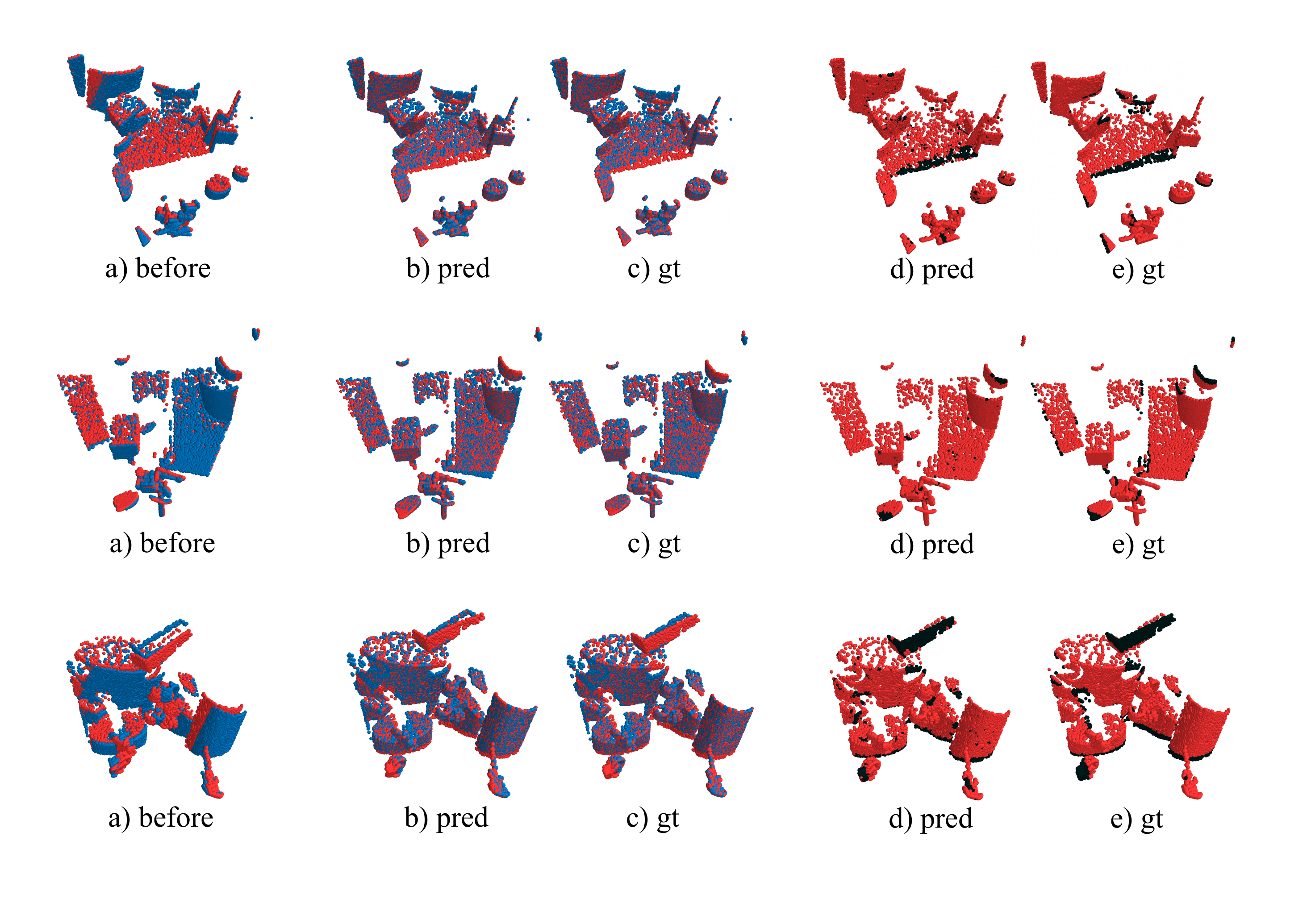}
\end{center}
\vspace{-55pt}
\caption{\textbf{Supervised visualization on Flyingthings3D}. We first train our model using the \textit{supervised} loss on the training set of Flyingthings3D and then show the prediction on 3 samples from the validation set. In a), we plot the source (red) and target (blue) point clouds from the dataset on the same 3D space. In b) and c), we show the warped source point cloud (source+flow) according to the predicted/ground truth scene and the target point cloud. The red isolated regions in these two plots represent the occlusion in the source since they do not have corresponding blue target regions. In d) and e), we show the predicted and ground truth occlusion map of the source, where the non-occluded regions are colored by red and the occluded regions are colored by black.}
\label{fig:ppwoc_sup_f3d}
\end{figure*}

\begin{figure*}
\begin{center}
\vspace{10pt}
\includegraphics[width=0.92\textwidth,keepaspectratio]{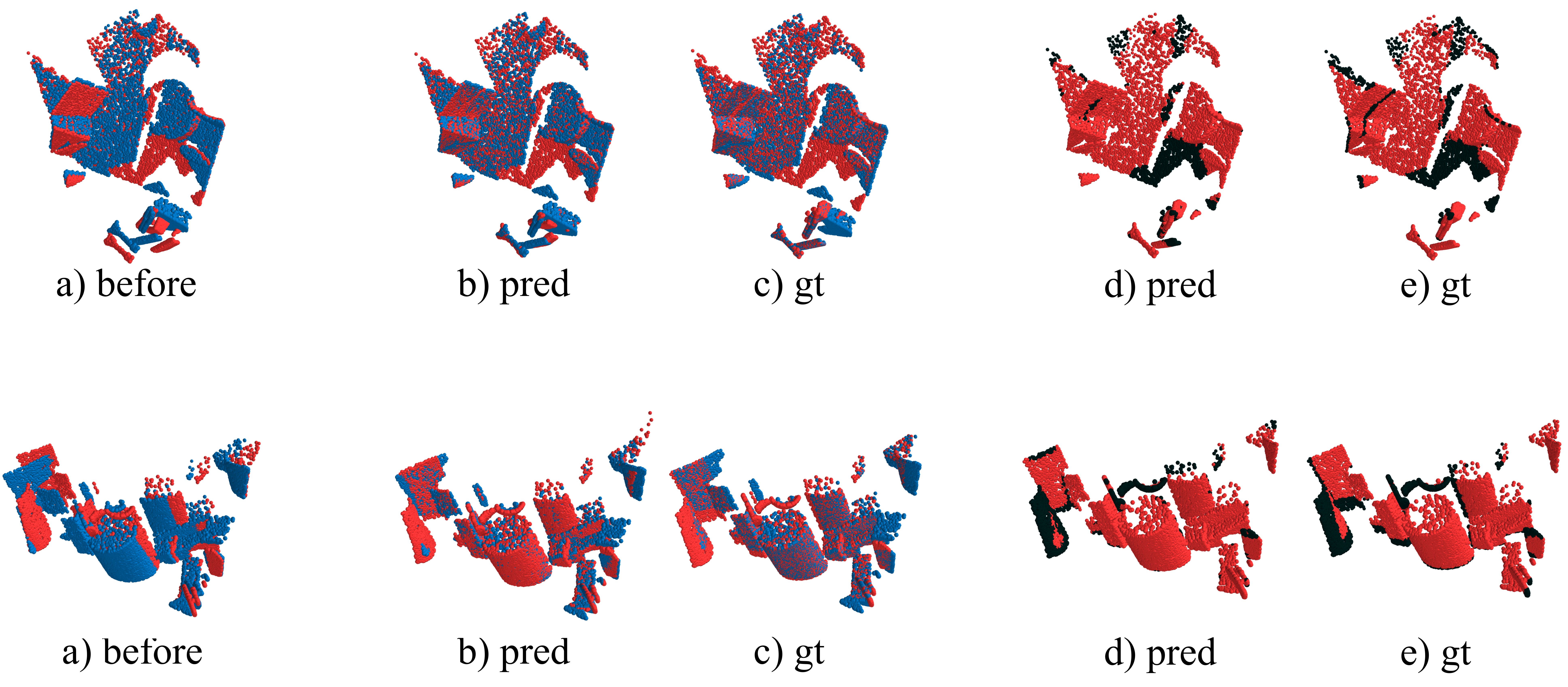}
\end{center}
\vspace{-22pt}
\caption{\textbf{Self-supervised visualization on Flyingthings3D}. 
We train our model using the \textit{Self-supervised} losses on the training set of Flyingthings3D and show the prediction on 2 samples from the validation set. In a), we plot the source (red) and target (blue) point clouds from the dataset on the same 3D space. In b) and c), we show the warped source point cloud (source+flow) according to the predicted/ground truth scene and the target point cloud. In d) and e), we show the predicted and ground truth occlusion map of the source, where the non-occluded regions are colored by red and the occluded regions are colored by black. As we can see, by using our novel self-supervised training scheme, both the scene flow and occlusion can be learned without any ground truth label.
}
\label{fig:ppwoc_self_f3d}
\end{figure*}

\begin{figure*}
\begin{center}
\includegraphics[width=1.0\textwidth]{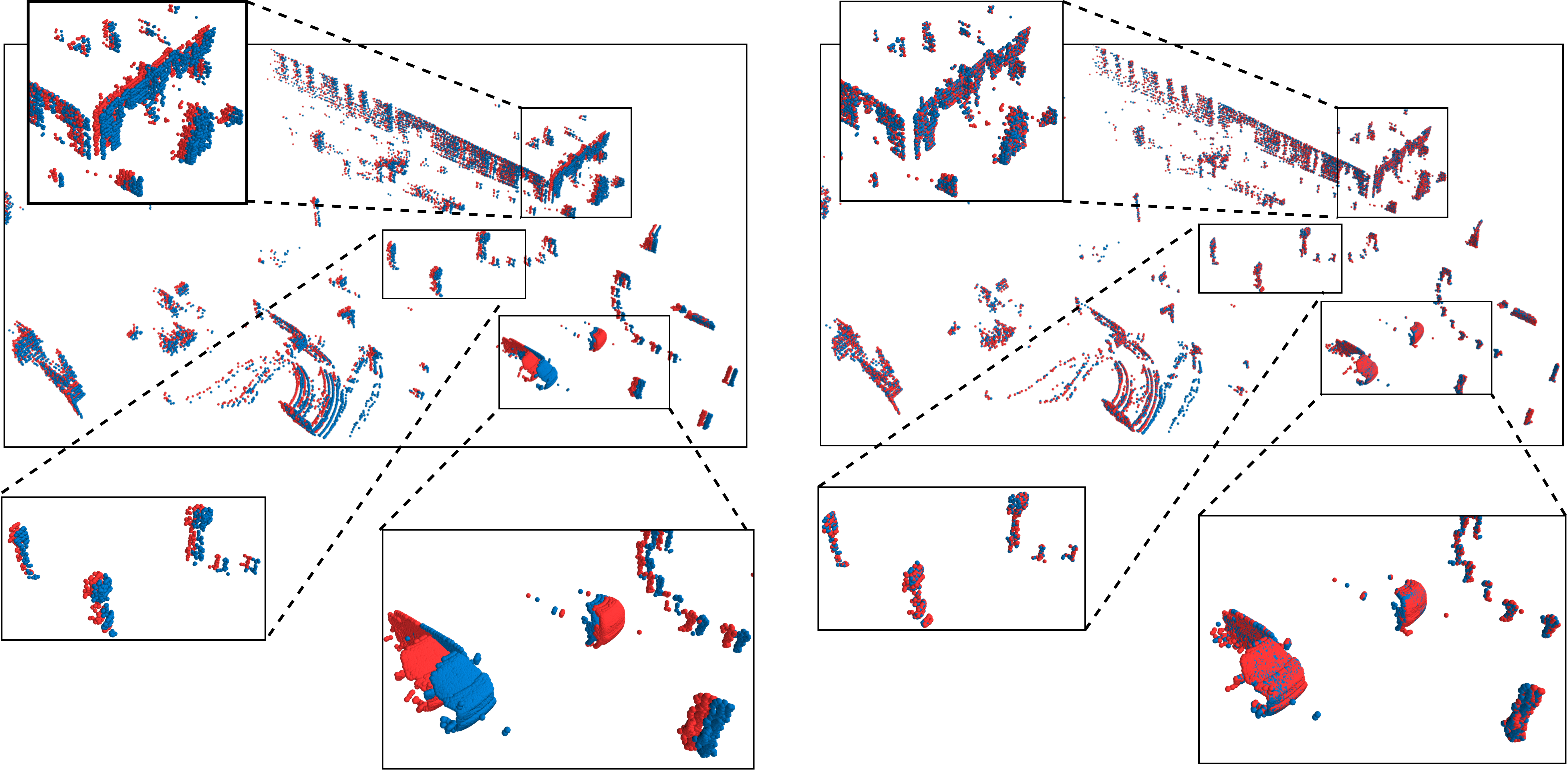}
\end{center}
\vspace{-15pt}
\caption{\textbf{Supervised visualization on KITTI}. On the left, we plot the source (red) and target (blue) point clouds from the KITTI dataset on the same 3D space. On the right, we plot the warped source according to the predicted scene flow and the target point cloud. The predicted scene flow is obtained from the \textit{supervised} pretrained model on Flyingthings3D without fine-tuning. The zoomed view for the marked regions is also provided. The better the alignment between the warped source and the target, the more accurate the estimated flow.}
\label{fig:ppwoc_sup_kitti}
\end{figure*}

\begin{figure*}
\begin{center}
\includegraphics[width=1.0\textwidth]{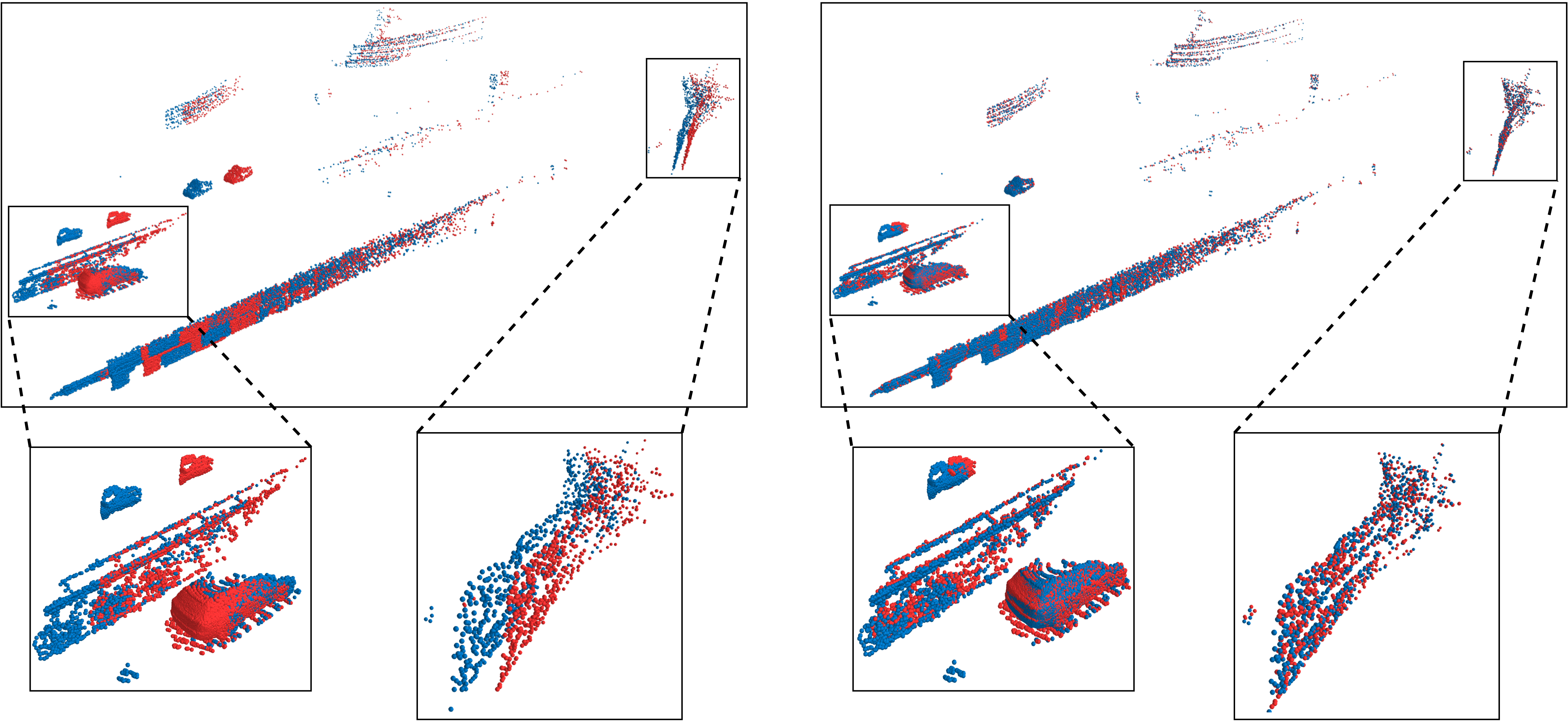}
\end{center}
\vspace{-15pt}
\caption{\textbf{Self-supervised visualization on KITTI}. On the left, we plot the source (red) and target (blue) point clouds of a sample from the test set of KITTI on the same 3D space. On the right, we plot the warped source according to the predicted scene flow and the target point cloud. The predicted scene flow is obtained from the \textit{self-supervised} pretrained + \textit{self-supervised fine-tuned} model as explained in the paper. The zoomed view for the marked regions is also provided. The better the alignment between the warped source and the target, the more accurate the estimated flow.}
\label{fig:ppwoc_self_kitti}
\end{figure*}

In figure~\ref{fig:ppwoc_sup_f3d}, we provide more visualization of the supervised training on the Flyingthings3D dataset. We first train our model using the supervised loss on the training set of the Flyingthings3D, and then we show the visualization of the predictions on the validation set. For better visualization of the predicted occlusion mask, we consider a point $s_i\in S$ to be occluded if its predicted occlusion probability is less than 0.5. A point is considered to be non-occluded if its occlusion probability is greater than or equal to 0.5. In figure~\ref{fig:ppwoc_self_f3d}, we provide more visualization of the self-supervised training on the Flyingthings3D dataset. As we can see, both the scene flow and occlusion label for the source point cloud can be learned accurately without any ground truth label by using our novel self-supervised learning scheme. In figure~\ref{fig:ppwoc_sup_kitti} and figure~\ref{fig:ppwoc_self_kitti}, we provide more visualization on the KITTI dataset. In figure~\ref{fig:ppwoc_sup_kitti}, we first train our model on the Flyingthings3D using the supervised loss, and then we show the predictions on the KITTI. In figure~\ref{fig:ppwoc_self_kitti}, we first train our model on the Flyingthings3D using the self-supervised loss, and then we perform the self-supervised fine-tuning on the training set of KITTI, a sample from the test set is shown.

\end{document}